\pdfoutput=1

\documentclass[11pt]{article}
\usepackage{mathtools}
\usepackage{algorithm}  
\usepackage{algpseudocode}
\usepackage[]{acl}
\usepackage{amsmath}
\usepackage{colortbl}
\usepackage{times}
\usepackage{bm}
\usepackage{latexsym}
\usepackage{booktabs, colortbl, xcolor, graphicx}
\usepackage{booktabs}
\usepackage{multirow}
\usepackage{amsfonts}
\usepackage{siunitx}
\usepackage{colortbl}
\usepackage{inconsolata}
\usepackage{microtype}
\usepackage{graphicx}
\usepackage{amssymb}
\usepackage{enumitem}

\usepackage[T1]{fontenc}

\usepackage[utf8]{inputenc}
\usepackage{microtype}

\usepackage{inconsolata}
\usepackage{pgfplots}      
\usepackage{caption}      
\usepackage{subcaption} 
\usepackage{adjustbox}
\title{Communication-Efficient and Tensorized Federated Fine-Tuning of Large Language Models}

\newcommand\crule[3][black]{\textcolor{#1}{\rule{#2}{#3}}}
\newcommand{\progressbarWidth}{3cm}
\newcommand{\progressbarHeight}{0.25cm}
\newcommand{\progressbar}[2]{\crule[#2]
            {\progressbarWidth*#1/100}
            {\progressbarHeight}\crule[gray!30]
            {\progressbarWidth*(100-#1)/100}
            {\progressbarHeight}}

\author{
        \textbf{Sajjad Ghiasvand}$^1$ \ \ 
        \textbf{Yifan Yang}$^2$ \ \ 
        \textbf{Zhiyu Xue}$^1$ \ \ \\
        \textbf{Mahnoosh Alizadeh}$^1$ \ \ 
        \textbf{Zheng Zhang}$^1$ \ \ 
        \textbf{Ramtin Pedarsani}$^1$
       \\ 
  Electrical and Computer Engineering Department, UC Santa Barbara$^1$ \ \ \\ Computer Science Department, UC Santa Barbara$^2$
  \\
  {\tt \{sajjad,yifanyang,zhiyuxue,alizadeh,zhengzhang,ramtin\}@ucsb.edu} 
}

\begin{document}
	\maketitle

\begin{abstract}
Parameter-efficient fine-tuning (PEFT) methods typically assume that Large Language Models (LLMs) are trained on data from a single device or client. However, real-world scenarios often require fine-tuning these models on private data distributed across multiple devices. Federated Learning (FL) offers an appealing solution by preserving user privacy, as sensitive data remains on local devices during training. Nonetheless, integrating parameter-efficient fine-tuning (PEFT) methods into FL introduces two main challenges: \textit{communication overhead} and \textit{data heterogeneity}.
In this paper, we introduce FedTT and FedTT+, methods for adapting LLMs by integrating tensorized adapters into client-side models' encoder/decoder blocks. FedTT is versatile and can be applied to both cross-silo FL and large-scale cross-device FL. FedTT+, an extension of FedTT tailored for cross-silo FL, enhances robustness against data heterogeneity by adaptively freezing portions of tensor factors, further reducing the number of trainable parameters. 
Experiments on BERT and LLaMA family models demonstrate that our proposed methods successfully address data heterogeneity challenges and perform on par or even better than existing federated PEFT approaches while achieving up to 10$\times$ reduction in communication cost.

\end{abstract}

\section{Introduction}
Large Language Models (LLMs) such as GPT-4~\cite{achiam2023gpt}, LLaMA~\cite{touvron2023llama}, and BERT~\cite{devlin2018bert} excel in tasks such as translation and summarization~\cite{bommasani2021opportunities} due to the capabilities of transformer architectures~\cite{vaswani2017attention}. While fine-tuning enhances adaptability~\cite{howard2018universal}, fully fine-tuning these massive models is computationally expensive and prone to overfitting. Parameter-efficient fine-tuning (PEFT) methods such as Adapters~\cite{houlsby2019parameter}, Prompt-Tuning~\cite{lester2021power}, and LoRA~\cite{hu2021lora} address this by optimizing only a subset of parameters, reducing costs without sacrificing performance~\cite{ding2023parameter}. However, traditional PEFT assumes centralized data, whereas real-world applications often involve private, distributed data, such as medical or legal records~\cite{manoel2023federated,shoham2023federated,soltanmohammadi2024optimizing}. Federated Learning (FL) emerges as a compelling solution, as it prioritizes user data privacy by ensuring sensitive information remains on individual devices during training. Instead of transmitting data to a central server, clients in FL locally update their model parameters and share only model information, such as parameters or gradients, which are then aggregated by the server into a global model~\cite{mcmahan2017communication}.

To address the aforementioned problems, FL has been integrated into PEFT methods
~\cite{zhang2023fedpetuning,fan2023fate,zhao2023fedprompt}.
These methods, however, often result in two major challenges: (i) communication and computation overhead in the FL system, and (ii) significant accuracy degradation, particularly under heterogeneous scenarios. 

Communication efficiency is crucial in FL, as edge devices typically have limited storage and computational power. While some methods attempt to reduce the number of training parameters by incorporating sparsity in PEFT approaches, they either perform poorly in FL~\cite{he2022sparseadapter,wu2022pruning} or are computationally expensive for clients~\cite{kuo2024federated}, rendering them unsuitable for practical scenarios.

Data heterogeneity happens when training data is not identically and independently distributed across clients (non-i.i.d.). In such scenarios, local models on individual clients can diverge from the global model's optimal state, leading to slower convergence~\cite{hsieh2020non,li2020federated}. This issue is particularly pronounced when training LLMs in federated environments, as data heterogeneity can severely impact model performance~\cite{zhang2023fedpetuning}. 
Although various studies have employed techniques such as gradient tracking to address this challenge in federated learning~\cite{ghiasvand2024robust,ebrahimi2024distributed}, these approaches often become inefficient in terms of communication and computational demands when applied to federated LLMs. 
Several works have modified LoRA to improve its efficiency in highly heterogeneous federated settings~\cite{babakniya2023slora}. However, these methods require communicating a large number of parameters (at least as much as LoRA). 

In this work, we present a federated and tensorized framework that reduces communication load compared to other PEFT methods, while maintaining comparable or even superior accuracy, and addressing challenges such as data heterogeneity. We introduce the Federated Tensor Train (FedTT) algorithm, where tensorized adapters serve as trainable parameters embedded within the encoder/decoder blocks of models used by clients. FedTT is applicable to both cross-silo FL, where all clients participate in training, and large-scale cross-device (LSCD) FL, where only a subset of clients is selected in each round. For cross-silo FL, we propose the FedTT+ algorithm, a heterogeneity-robust method that further reduces the number of parameters by adaptively freezing portions of the tensor factors in the models. We summarize our contributions as follows: 
\begin{itemize} [leftmargin=*]
\vspace{-5pt}
\item We study the TT decomposition of adapters to enable communication-efficient federated fine-tuning of LLMs. 
\vspace{-5pt}
\item We propose FedTT, an efficient FL method to fine-tune LLMs using tensorized adapters, achieving up to $10\times$ communication reduction than other popular federated PEFT methods. 


\vspace{-5pt}
\item We introduce FedTT+, an enhancement of FedTT that further reduces the number of trainable parameters and improves robustness against data heterogeneity in cross-silo FL. FedTT+ outperforms state-of-the-art (SOTA) cross-silo FL methods with fewer trainable parameters. 

\vspace{-5pt}
\item We conduct extensive experiments across various settings, including data heterogeneity and differential privacy, using widely adopted LLMs such as the BERT and LLaMA-2 model families. These experiments validate the effectiveness of our proposed algorithms in reducing communication overhead while preserving high accuracy.
\end{itemize}

\section{Related Work}

\subsection{Parameter Efficient Fine-Tuning (PEFT)}
PEFT methods can be generally divided into three categories~\cite{han2024parameter}. The first is Additive PEFT, where a small set of trainable parameters is added to the model, and only these are updated during training. Methods like
Prefix-tuning~\cite{li2021prefix} and Prompt-tuning~\cite{lester2021power} fall into this category, and our approach aligns with this strategy. The second is Selective PEFT, which chooses a subset of existing model parameters for tuning, as seen in techniques like BitFit~\cite{zaken2021bitfit} and PaFi~\cite{liao2023parameter}. Lastly, Reparameterized PEFT introduces a low-rank parameterization of pre-trained weights for training, with methods such as LoRA~\cite{hu2021lora} and DoRA~\cite{liu2024dora}.


    

\subsection{PEFT in Federated Setting}\label{FedPEFT}
\citep{zhang2023fedpetuning} tests and compares various PEFT methods such as Adapter, LoRA, Prompt Tuning, and BitFit in FL. 
Several works have modified LoRA to improve its efficiency in highly heterogeneous federated settings. For example, SLoRA~\cite{babakniya2023slora,yan2024federa} modify initialization to address data heterogeneity, whereas HetLoRA~\cite{cho2023heterogeneous} and FlexLoRA~\cite{bai2024federated} adaptively adjust LoRA ranks for each client to handle system heterogeneity. 
However, these methods still require communicating massive parameters. 

Recently, \cite{kuo2024federated} introduced sparse fine-tuning to reduce the communication load in federated LoRA. Although this method lowers communication overhead and shows robustness to data heterogeneity in some tasks, it suffers from computational inefficiency due to the extensive matrix computations required during each communication round, both on the server and client sides.
FFA-LoRA~\cite{sun2024improving} and RoLoRA~\cite{chenrobust} aim to improve accuracy in the presence of heterogeneity while simultaneously reducing trainable parameters. Our algorithms can reduce communication overhead while achieving comparable or better accuracy to these approaches, particularly when data heterogeneity exists. 

\cite{kim2023client} employs hypernetworks for adapters to reduce the number of trainable parameters, and this approach can be integrated with any PEFT method, including ours. However, it introduces additional computational overhead, as each client must generate adapters from the hypernetwork~\cite{hu2024federated}.


\subsection{Tensor-based Model Compression}
Tensor compression has shown its great potential to reduce model size and enhance training efficiency. Early work by \citet{novikov2015tensorizing} employeed the TT format for network compression. Recent efforts in tensorized fine-tuning of LLMs have shown promise in achieving high performance with significantly fewer parameters compared to traditional fine-tuning methods. However, these approaches still involve training a large number of variables, especially when compared to the more recent methods like LoRA. In response, \citet{yang2024loretta} developed the Low-Rank Economic Tensor-Train Adaptation, which innovates by using tensorized-layer based adapters and reshaping update matrices into smaller tensor factors. Despite demonstrating substantial reductions in trainable parameters—up to 100 times less than popular PEFT methods—the performance of these tensorized approaches in FL scenarios, particularly in the presence of data heterogeneity, remains an open question.


\section{Preliminaries  }

     \begin{figure*}[t]
     \vspace{-15pt}
  \centering
\includegraphics[width=0.85\textwidth]{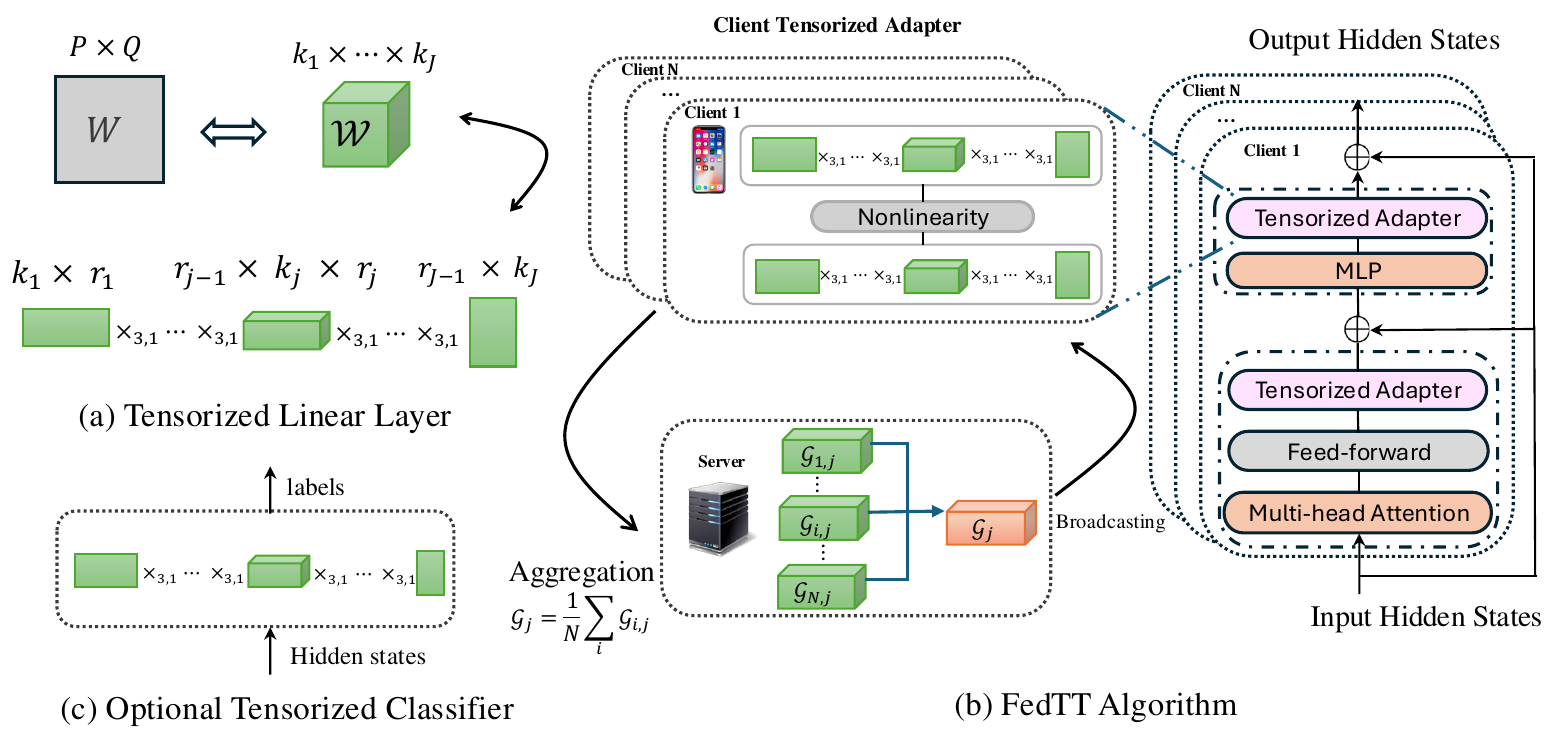}
  \caption{Illustration for the tensorized linear layer (a), FedTT algorithm (b), and the optional tensorized classifier applied for classification tasks (c). The FedTT algorithm workflow includes fine-tuning clients' tensorized adapters, aggregating tensor factors on the server, and broadcasting the updated weights back to clients. 
  }
  \label{fig:method_adapter}
  \vspace{-10pt}
\end{figure*}

\subsection{Federated Fine-tuning}
Federated fine-tuning is a distributed approach for collaboratively fine-tuning a global model across a central server and a network of $N$ clients, denoted by $ \mathcal{C} = \{c_1,\ldots,c_N\} $. The objective is to optimize the global trainable parameters $\bm{w}$ by minimizing the following objective function: 
\begin{align*} \min_{\bm{w} }\mathcal{L}(\tilde{\bm{w}},\bm{w}) = \frac{1}{N}\sum_{i=1}^N \ell_i(\tilde{\bm{w}},\bm{w};\mathcal{D}_i), \end{align*} 
where $\tilde{\bm{w}}$ is the pre-trained model parameters, which are identical and fixed for all clients, $ \ell_i(\cdot) $ represents the local objective, and $ \mathcal{D}_i $ represents the local data distribution for client $c_i$.

A well-known FL method is the FedAvg algorithm~\cite{mcmahan2017communication}. The server selects a subset of clients $ \mathcal{S} \subseteq \mathcal{C} $ in each training round. Each chosen client $ c_s \in \mathcal{S} $ initializes its local model with the global model parameters from the previous round $ \bm{w}^{(t)} $, and then performs local training using stochastic gradient descent on its dataset for $ K $ local updates: 
\begin{equation}
\begin{aligned}
    \bm{w}_s^{(t)+k+1} \longleftarrow \bm{w}_s^{(t)+k} - \eta \nabla \ell_s(\tilde{\bm{w}},\bm{w}_s^{(t)+k};\mathcal{D}_s),\nonumber
\end{aligned}
\end{equation}
where $ \eta $ is the learning rate and $ \bm{w}_s^{(t)+k} $ refers to the local model parameters for client $ c_s $ during communication round $ t $ and local update $ k $. After local training is completed, clients send their updated model parameters to the server, which then aggregates these updates to form the new global model: 
$
\boldsymbol{w}^{(t+1)}=1/|\mathcal{S}| \sum_{s=1}^{|\mathcal{S}|} \boldsymbol{w}_s^{(t)+K}.\nonumber
$

Despite its advantages, FL faces two major challenges: (i) the large size of local models, which results in significant communication overhead, and (ii) data heterogeneity, which can cause local models to diverge from each other.

\subsection{Tensor-Train Decomposition}\label{sec:tensor}

In this subsection, we provide some introduction to the tensor and tensor train (TT) decomposition \cite{oseledets2011tensor}. Tensors are natural multidimensional generalizations of matrices. The tensor $\mathcal{W} \in \mathbb{R}^{k_1 \times \cdots \times k_J}$ is indexed as $\mathcal{W}=(w_{i_1\cdots i_J})_{1\leq i_j\leq k_j}$ said to have order of $J$ and dimension $k_1, \cdots, k_J$. Given two tensors $\mathcal{W} \in \mathbb{R}^{k_1 \times \cdots \times k_J}$ and $\mathcal{V} \in \mathbb{R}^{l_1 \times \cdots \times l_M}$ with $k_s=l_t$, the multiplication between two tensors $\mathcal{C} = \mathcal{W} \times_{s,t} \mathcal{V}$ can be performed as:
\begin{align*}
    \mathcal{C}_{(i_p)_{p\neq s}, (j_p)_{p\neq t}} = \sum_{i_s=j_t=1}^{k_s} w_{i_1\cdots i_s \cdots i_J} v_{j_1\cdots j_t \cdots j_M}.
\end{align*}
The TT decomposition serves as a potent alternative to traditional matrix decomposition techniques, which decompose a large tensor into a list of tensor factors \cite{oseledets2011tensor} by TT-SVD method. As shown in Fig. \ref{fig:method_adapter} (a), to decompose the weight matrix into small tensors, we begin by reshaping a matrix $\boldsymbol{W} \in \mathbb{R}^{P \times Q}$ into a  tensor $\mathcal{W} \in \mathbb{R}^{k_1 \times \cdots \times k_J}$. Then $\mathcal{W}$ can be parameterized compactly via a sequence of $J$ tensor factors $\mathcal{G}_1, \ldots, \mathcal{G}_J$ as: 
 \begin{align}
 \label{eq:TT}
    \mathcal{W} = \mathcal{G}_1 \times_{3,1} \mathcal{G}_2\times_{3,1} \cdots\times_{3,1} \mathcal{G}_J,
\end{align}
Here, each tensor factor $\mathcal{G}_j$ has the shape of $\mathcal{G}_j \in \mathbb{R}^{r_{j-1} \times k_j \times r_j}$, where $\boldsymbol{r}=(r_0,r_1, \cdots, r_J)$ is the tensor rank, and the product of dimensions $\Pi_j k_j = P \cdot Q$. The setup of tensor ranks follows the boundary conditions with $r_0 = r_J = 1$, while the other ranks $r_j, j \notin {0, J}$ are chosen based on specific tasks or made adaptive \cite{yang2024comera}. 

As we can see, the tensorized layer substantially reduces the parameter count for the weight matrix \(\bm{W}\) from \(P\times Q\) to \(\sum_{i=1}^d r_{i-1}k_i r_i\), offering a much higher compression ratio than previous PEFT work. Unlike the traditional Adapters method \cite{houlsby2019parameter}, which uses a bottleneck structure to reduce trainable parameters, our tensorized adapters achieve an even larger compression ratio by using two tensorized linear layers with a nonlinear activation in between. For example, with weight $\boldsymbol{W}$ with size $768\times 768$ and bottleneck size \(64\), a standard Adapter incurs $2\cdot 768 \cdot 64\approx98K$
trainable parameters for its weight matrices, whereas our method adds only $\sum_{i=1}^{6}(5^2\cdot 8)=1.2K$ parameters (assuming core dimensions \([8,8,8,8,8,8]\) for $\mathcal{W}$ and TT rank 5). This high compression ratio enables FedTT to achieve higher performance under a similar communication cost compared with other PEFT approaches.

Instead of performing TT decomposition on the weight matrix, we directly initialize, store, and update the list of tensor factors in this work, as shown in Fig. \ref{fig:method_adapter} (a). During the forward pass, the tensor factors are directly contracted with the vector of activation values, and the weight matrix $\boldsymbol{W} $ does not need to be reconstructed.  Since the size of the tensor factors is small, the contraction process is significantly faster than the original matrix-vector product~\cite{yang2024comera}.

\section{Proposed Algorithm}

In this section, we introduce the proposed FedTT method, which is built on a novel parameter-efficient tensorized adapter. This adapter is integrated into clients' local models to effectively adapt LLMs. We begin with discussing the structure and setup of the tensorized adapters, followed by the introduction of FedTT and its enhanced version, FedTT+, which can be selected based on communication budget constraints. The workflow of our FedTT method is shown in Fig. \ref{fig:method_adapter} (b).
\subsection{Tensorized Adapters}\label{sec:tensor_adapters}
To facilitate fine-tuning in a FL framework, we incorporate tensorized adapters designed to adapt LLMs with minimal additional parameters. These adapters are underpinned by the novel tensorized linear layer \cite{yang2024loretta, yang2024adazeta}, which replaces the matrix weight in linear layers with more parameter-efficient TT format weights. This section first details the architecture of the tensorized linear layer and then elucidates the process of constructing tensorized adapters. 

In Sec. \ref{sec:tensor}, we introduced how to represent a weight matrix in the TT format, where the TT format tensor factors $\mathcal{G}_1, \ldots, \mathcal{G}_J$ are initialized, stored, and updated during the fine-tuning process. The tensorized layer is successfully integrated into matrix-based LLMs, performing tensor contraction over the list of tensor factors and reshaping them to match the traditional weight matrix $\boldsymbol{W}$. Compared to traditional matrix-based linear layers, the tensorized layers only store the smaller tensor factors, significantly reducing the number of parameters. 

We now introduce the tensorized adapters, built based on the tensorized layers in Fig. \ref{fig:method_adapter}. These adapters use a bottleneck structure similar to that in the original sequential adapter methodology \cite{houlsby2019parameter}, comprising two tensorized linear layers with a nonlinear layer in between. The bottleneck structure further reduces the number of trainable parameters by decreasing the channels connected between the two tensorized layers and the nonlinearity layer, from the size of the hidden dimension to a smaller number, such as 64. As illustrated in Fig. \ref{fig:method_adapter} (b), the tensorized adapter is strategically placed after the attention and MLP components of an encoder/decoder block. In practical applications, we compress the classification head (linear layer) into a tensorized layer and make it trainable for sequence classification tasks, as shown in Fig. \ref{fig:method_adapter} (c). Instead, we retain the original language model head without compression for general language modeling tasks, as experiments show a significant reduction in performance when the language modeling head is compressed.
\subsection{FedTT Method}

\begin{algorithm}[t!]
	\caption{FedTT} \label{al.1}
 {\small
	\begin{algorithmic}[1]
            \For {communication round $ t\leftarrow1 $ to $ T $}
		\For {clients $ c_i \in \mathcal{C} $ in parallel} 
			
                        \For 
                        {local update $ k\leftarrow1 $ to $ K $}
                        \State $\bm{w}_i^{(t)+k+1} \leftarrow \text{Update}(\bm{w}_i^{(t)+k},\mathcal{D}_i)$
                        \EndFor
                        \State Client $c_i$ sends $ \bm{w}_i^{(t)+K} $ to the server
                        \EndFor
                        \State At server: $\bm{w}^{(t+1)} = \frac{1}{N}\sum_{i=1}^N \bm{w}_i^{(t)+K}$
                        \State Server sends $\bm{w}^{(t+1)}$ to all clients in $\mathcal{C}$         
				\EndFor
	\end{algorithmic} 
 }
\end{algorithm}

We introduce the FedTT algorithm, as outlined in Alg.~\ref{al.1}. In FedTT, the full model weights and architecture are initially distributed to all clients in the set $\mathcal{C}=\{c_1, \cdots, c_N\}$ at the start of the fine-tuning process. Tensorized adapters are then injected into each client's local model and designated as the trainable parameters. FedTT operates over $T$ communication rounds, where each client performs $K$ local updates on its trainable parameters during each round.

During the communication round $t$, each client $c_i \in \mathcal{C}$ updates its trainable parameters $\bm{w}^{(t)}_i$, including the tensorized classifier, tensorized adapter, and bias terms (if they exist), for $K$ local updates. Afterward, each client sends its updated parameters, $\bm{w}^{(t)+K}_i$, to the central server. The server then aggregates the weights as
$\bm{w}^{(t+1)} = 1/N\sum_{i=1}^N \bm{w}^{(t)+K}_i$
for the next training round and sends updated weights $\bm{w}^{(t+1)}$ back to the clients. Note that while we have primarily described cross-silo FL scenarios, the LSCD FL setting operates similarly, with the key distinction being that in each communication round, the server sends the aggregated weights to a randomly chosen subset of clients  $\mathcal{S} \subseteq \mathcal{C}$, rather than to all clients in $\mathcal{C}$.

Compared with most previous federated PEFT methods, FedTT significantly reduces communication overhead by only transferring small tensor factors. This approach leads to over $10\times$ communication reduction compared to LoRA adapters, as demonstrated in Sec. \ref{sec:tensor_adapters}, making FedTT particularly advantageous in FL scenarios where efficient communication is critical.
\subsection{FedTT+ Method}
\begin{algorithm}[t!]
	\caption{FedTT+} \label{al.2}
 {\small
	\begin{algorithmic}[1]
			\For {communication round $ t\leftarrow1 $ to $ T $}
   \For {clients $ c_i \in \mathcal{C} $ in parallel}
   \State $r \leftarrow$ (choose index $r$ s.t. \text{mod}$(t, {\scriptsize J-2})$ $=$ $r-1$)
                        \For 
                        {local update $ k\leftarrow1 $ to  $ K $}
                        \vspace{5pt}
                        \State  for   $ h\in\{1,r,J\}$:\\ \quad\quad\quad\quad\quad\quad$\mathcal{G}_{i,h}^{(t)+k+1} \leftarrow \text{Update}(\mathcal{G}_{i,h}^{(t)+k}; \mathcal{D}_i)$
                        \vspace{1pt}
                        \EndFor
                        \State Client $c_i$ sends {\footnotesize$\{\mathcal{G}_{i,1}^{(t)+K},\mathcal{G}_{i,r}^{(t)+K},\mathcal{G}_{i,J}^{(t)+K}\}$} to  server
                        
				\EndFor
			
  \State At server: {\footnotesize $\mathcal{G}^{(t+1)}_h = \frac{1}{N}\sum_{i=1}^N \mathcal{G}_{i,h}^{(t)+K}$} for $ h\in\{1,r,J\}$
                        \State Server sends {\footnotesize $\{\mathcal{G}_1^{(t+1)},\mathcal{G}_r^{(t+1)},\mathcal{G}_J^{(t+1)}\}$} to all clients in $\mathcal{C}$
  \EndFor
	\end{algorithmic} 
 }
\end{algorithm}
In this section, we propose an improved version of the FedTT method, named FedTT+ to further reduce the number of trainable parameters and enhance FedTT's suitability for scenarios with data heterogeneity. Before introducing FedTT+, we first explain the intuitive idea behind it.

In FedTT, the loss for back-propagation is computed on the product of tensor factors $\mathcal{G}_1, \ldots, \mathcal{G}_J$. However, under the federated setup, the aggregations in the server are performed separately on $\mathcal{G}_1, \ldots, \mathcal{G}_J$. This practice introduces additional terms in the product of the averaged $\mathcal{G}_1, \ldots, \mathcal{G}_J$, which may slow down the convergence of the algorithms. Ideally, the aggregation on the server should be performed on the product of the low-rank matrices $\mathcal{G}_1, \ldots, \mathcal{G}_J$. The left-hand side of Eq. \eqref{eq.1} shows the parameters after aggregation with FedTT using FedAvg, while the right-hand side represents the ideal aggregation.
\begin{equation}
\footnotesize
\begin{aligned}
\left(\frac{1}{N}\sum_{i=1}^N\mathcal{G}_{i,1} \right)&\times_{3,1}\ldots\times_{3,1}\left(\frac{1}{N}\sum_{i=1}^N\mathcal{G}_{i,J}\right)
\\&\neq \frac{1}{N}\sum_{i=1}^N \mathcal{G}_{i,1}\times_{3,1}\ldots\times_{3,1}\mathcal{G}_{i,J},
\end{aligned}\label{eq.1} 
\normalsize
\end{equation}
where $\mathcal{G}_{i,j}$ is the tensor factor $j$ for clientuser $c_i$.

\begin{table*}[t]
    \centering
    \caption{Comparative analysis of various federated PEFT methods using the DeBERTa-Base model in a cross-silo FL setting with an i.i.d. data distribution and $5$ clients.}
    \small
\resizebox{0.85\textwidth}{!}{%
    \begin{tabular}{l|c|ccccc|lrl}
    \toprule
    Model \& Method & \# Param.  & MRPC & SST-2 & QNLI & QQP & MNLI & Avg. \\
    \midrule
    DeBERTa-Base (LoRA$_{r=8}$) & $0.30M$ & $91.87$  &  $94.95$ & $92.68$  & $89.2$ & $87.31$ & $91.10$ \\
    DeBERTa-Base (P-Tuning) & $0.30M$ & $82.01$ &  $90.48$ &  $82.12$ & $84.0$ & $80.74$ & $83.87$  \\
    \midrule
    DeBERTa-Base (LoRA$_{r=4}$)  & $0.15M$ & $91.72$ & $\bm{94.95}$  & $\bm{92.66}$ & $86.7$ & $\bm{86.91}$ & $90.58$ \\
    
    DeBERTa-Base (BitFit) & $0.10M$ & $91.33$ & $94.72$ &  $91.89$ & $88.4$ & $86.02$ & $90.47$  \\

    DeBERTa-Base (RoLoRA$_{r=4}$) & $0.08M$ & $91.17$ & $94.61$ &  $92.40$ & $87.9$ & $86.27$ & $90.47$  \\
    DeBERTa-Base (Prompt) & $\bm{0.01M}$ & $82.96$ & $92.32$ &  $82.13$ & $80.5$ & $74.46$ & $82.47$  \\
    \rowcolor{blue!10} 
    \textbf{DeBERTa-Base (FedTT)} & $0.06M$ & $\bm{92.68}$  &  $94.61$ & $92.02$  & $\bm{88.4}$ & $85.99$ & $\bm{90.74}$ \\
     \rowcolor{blue!10} 
    \textbf{DeBERTa-Base (FedTT+)} & $\bm{0.02M}$ & $92.60$  &  $93.58$ & $90.54$  & $87.9$ & $85.33$ & $89.99$ \\

    \bottomrule
    \end{tabular}}
\label{table1}
\end{table*}

\noindent
It is important to note that the difference between the right-hand side and the left-hand side of Eq. \eqref{eq.1} becomes more significant when:  
i) the number of clients is large,  
ii) the clients have non-IID data distributions, and  
iii) the number of local updates in each communication round is substantial.

In FedTT+, we alleviate this interference problem by freezing \underline{most of the tensor factors} in each communication round and only updating \underline{a small fraction of them}. In this approach, most of the tensor factors remain fixed and identical across all clients during training while only a few tensor factors are trainable in each communication round. For example, assume that we just update $ \mathcal{G}_1 $ and freeze $\mathcal{G}_2, \ldots, \mathcal{G}_J$ in communication $ t $. Then, $\mathcal{G}_2, \ldots, \mathcal{G}_J$ are identical across the clients and Eq. \eqref{eq.1} can be re-written as 
\begin{equation}
\footnotesize
\begin{aligned}
\label{eq.2}
&\left(\frac{1}{N}\sum_{i=1}^N\mathcal{G}^{(t)}_{i,1}\right)\times_{3,1} \mathcal{G}^{(t-1)}_{2}\times_{3,1}\ldots\times_{3,1}\mathcal{G}_{J}^{(t-1)} \nonumber\\ &=\frac{1}{N}\sum_{i=1}^N \left(\mathcal{G}^{(t)}_{i,1}\times_{3,1}\mathcal{G}^{(t-1)}_{2}\times_{3,1}\ldots\times_{3,1}\mathcal{G}_{J}^{(t-1)}\right).\nonumber
\end{aligned} 
\end{equation}
This modification to FedTT can improve accuracy, particularly in cases of severe data heterogeneity. The detailed algorithm is presented in Algorithm~\ref{al.2}.

FedTT+ operates similarly to FedTT, but with a key difference: in each communication round $t$, an index $r$ is selected from the set $ r \in \{2,\ldots,J-1\} $ (line 3 in Alg.~\ref{al.2}). In this process, $ \{\mathcal{G}_1, \mathcal{G}_r, \mathcal{G}_J \}$ are set as trainable parameters (as the first and last tensors are always trained), while the other factors, $ \{\mathcal{G}_2, \ldots, \mathcal{G}_{r-1}, \mathcal{G}_{r+1}, \ldots, \mathcal{G}_{J-1}\} $, remain frozen (line 5 and 6 in Alg.~\ref{al.2}). As a result, clients only send their updated $ \{\mathcal{G}_{i,1}, \mathcal{G}_{i,r}, \mathcal{G}_{i,J} \}$ to the server for aggregation, significantly reducing communication overhead. Note that the classification head is always trainable and present in all communication rounds.

\cite{sun2024improving,chenrobust} have also demonstrated that freezing certain parameters improves the performance of LoRA in the presence of data heterogeneity, particularly with larger models like RoBERTa-large. In the numerical section, we compare FedTT+ with their method under the same settings and show that our approach achieves comparable or better accuracy while using even fewer trainable parameters.




\section{Numerical Results}
\subsection{Experiments Setup}
\begin{table*}[t]
    \centering
    \caption{Comparative analysis of various federated PEFT methods using the RoBERTa-Base model. The reported accuracy for federated LoRA, Adapter, Prefix, and BitFit methods is sourced from~\cite{zhang2023fedpetuning}. }
    \small
\resizebox{0.85\textwidth}{!}{%
    \begin{tabular}{l|c|ccccc|lrl}
    \toprule
    Model \& Method & \# Param. & MRPC & SST-2 & QNLI & QQP & MNLI & Avg. \\
    \midrule
    RoBERTa-Base (Prefix) & $3.50M$ & $88.1$ & $93.7$ &  $84.6$ & $81.8$ & $80.4$ & $85.7$  \\
    RoBERTa-Base (P-Tuning) & $0.88M$ & $86.8$ &  $92.1$ &  $85.6$ & $81.5$ & $79.8$ & $85.2$  \\
    RoBERTa-Base (Adapter) & $0.70M$ & $88.5$ &  $94.0$ &  $85.9$ & $\bm{87.0}$ & $\bm{84.9}$ & $88.1$  \\
    RoBERTa-Base (IA3) & $0.65M$ & $88.0$ & $93.0$ &  $\bm{89.4}$ & $85.4$ & $82.7$ & $87.7$  \\
    RoBERTa-Base (LoRA) & $0.30M$ & $\bm{89.8}$  &  $\bm{94.4}$ & $86.0$  & $86.5$ & $84.7$ & $88.3$ \\

    RoBERTa-Base (BitFit) & $0.10M$ & $88.6$ & $92.8$ &  $80.5$ & $84.0$ & $80.7$ & $85.3$  \\
    \rowcolor{blue!10} 
    \textbf{RoBERTa-Base (FedTT)} & $\bm{0.06M}$ & $88.9$  &  $93.8$ & $88.9$  & $86.2$ & $84.2$ & $\bm{88.4}$ \\
    \bottomrule
    \end{tabular}}
\label{table2}
\end{table*}

We conduct extensive experiments to evaluate the performance of the proposed algorithms across various language models. For the BERT-family models, we utiliz RoBERTa-base~\cite{liu2019roberta}, DeBERTa-base~\cite{he2020deberta}, and RoBERTa-large~\cite{liu2019roberta}, while for large-scale models, we employed LLaMA-2~\cite{touvron2023llama}. Using these models, we compare the proposed algorithm against several PEFT methods in FL scenarios, including BitFit, LoRA, Adapter, Prefix-Tuning, and Prompt-Tuning. Additionally, we benchmark it against SOTA federated PEFT methods such as FFA-LoRA~\cite{sun2024improving} and RoLoRA~\cite{chenrobust}.

We consider two main FL scenarios: the cross-silo FL scenario~\cite{kairouz2021advances} and the LSCD FL scenario~\cite{lai2022fedscale}. 
The cross-silo scenario is suitable for networks with typically fewer than $100$ clients. In this case, the server sends the updated model to all clients, meaning every client participates in training during each communication round. In contrast, LSCD FL is more appropriate for environments with thousands of clients, where only a randomly selected subset of clients is involved in each training round. 
For cross-silo FL, the number of clients is set to $n \in \{5, 10, 20, 50\}$, while for LSCD FL, similar to~\cite{zhang2023fedpetuning}, we randomly select $10$ clients from a pool of $1000$. All experiments are conducted using the AdamW optimizer~\cite{loshchilov2018decoupled}, with a similar learning rate and batch size setup across different methods. We perform the experiments on NVIDIA A6000 and V100 GPUs.

\subsection{Performance on the BERT Family}

\begin{table*}[t]
    \centering
    \caption{
Comparison of SOTA cross-silo FL methods using RoBERTa-Large models under varying degrees of data heterogeneity. The accuracies for LoRA, FFA-LoRA, and RoLoRA methods are sourced from~\cite{chenrobust}.}
    \small
    	\resizebox{0.85\textwidth}{!}{%
    \begin{tabular}{c|l|c|cccc|lrl}
    \toprule
    Data Dist. & Model \& Method  & \# Param. &  SST-2 & QNLI & QQP & MNLI & Avg. \\
    \midrule
    \multirow{5}{*}{ i.i.d. } & RoBERTa-Large (LoRA$_{r=2}$) & $68K$ &  $\bm{95.64}$ & $92.04$  & $85.85$ & $86.16$ & $89.92$ \\
    &RoBERTa-Large (FFA-LoRA$_{r=2}$) & $34K$ &  $94.91$ & $90.11$  & $84.06$ & $85.48$ & $88.64$ \\
    &RoBERTa-Large (RoLoRA$_{r=2}$) & $34K$  &  $95.60$ & $91.62$  & $85.66$ & $86.16$ & $89.76$ \\
    &RoBERTa-Large (FedTT) & $51K$  &  $94.38$ & $93.01$  & $88.30$ & $87.20$ & $90.72$ \\
    &\textbf{RoBERTa-Large (FedTT+)}\cellcolor{blue!10} & $\bm{28K}$\cellcolor{blue!10} &  $\bm{95.64}$ \cellcolor{blue!10}& $\bm{94.05}$  \cellcolor{blue!10}& $\bm{88.99}$ \cellcolor{blue!10}& $\bm{88.27}$ \cellcolor{blue!10}& $\bm{91.74}$ \cellcolor{blue!10}\\
    \midrule
    \multirow{5}{*}{ mild het. } & RoBERTa-Large (LoRA$_{r=2}$) &  $68K$ &  $94.27$ & $86.91$  & $81.22$ & $82.07$ & $86.12$ \\
    &RoBERTa-Large (FFA-LoRA$_{r=2}$) & $34K$ &  $93.92$ & $89.58$  & $80.51$ & $82.62$ & $86.66$ \\
    &RoBERTa-Large (RoLoRA$_{r=2}$) &  $34K$ &  $94.84$ & $90.77$  & $85.13$ & $85.10$ & $88.96$ \\
    &RoBERTa-Large (FedTT) &  $51K$ &  $94.15$ & $91.38$  & $86.25$ & $86.53$ & $89.58$ \\
    &\textbf{RoBERTa-Large (FedTT+)}\cellcolor{blue!10} & $\bm{28K}$\cellcolor{blue!10} &  $\bm{95.64}$\cellcolor{blue!10} & $\bm{92.60}$\cellcolor{blue!10}  & $\bm{87.76}$ \cellcolor{blue!10}& $\bm{88.11}$\cellcolor{blue!10} & $\bm{91.03}$
    \cellcolor{blue!10} \\
    \midrule
    \multirow{5}{*}{ sever het. }  & RoBERTa-Large (LoRA$_{r=2}$) &  $68K$ &  $93.23$ & $82.57$  & $58.96$ & $76.96$ & $77.93$ \\
    &RoBERTa-Large (FFA-LoRA$_{r=2}$) & $34K$  &  $92.32$ & $85.15$  & $62.79$ & $77.78$ & $79.51$ \\
    &RoBERTa-Large (RoLoRA$_{r=2}$)&  $34K$ &  $\bm{94.61}$ & $89.83$  & $85.15$ & $85.55$ & $88.78$ \\
    &RoBERTa-Large (FedTT)&  $51K$ &  $94.38$ & $\bm{90.55}$  & $85.47$ & $85.27$ & $88.92$ \\
    &\textbf{RoBERTa-Large (FedTT+)}\cellcolor{blue!10} & $\bm{28K}$\cellcolor{blue!10}  &  $94.50$\cellcolor{blue!10} & $90.17$ \cellcolor{blue!10} & $\bm{86.65}$\cellcolor{blue!10} & $\bm{86.28}$\cellcolor{blue!10} & $\bm{89.40}$\cellcolor{blue!10} \\
    \bottomrule
    \end{tabular}}
\label{table3}
\end{table*}

\begin{table}[t]
    \centering
    \caption{Comparison of cross-silo FL methods using RoBERTa-Large models under varying differential privacy guarantees. The accuracies for LoRA and FFA-LoRA methods are sourced from~\cite{sun2024improving}.}\label{tab:priv}
    \small
\resizebox{\linewidth}{!}{%
    \begin{tabular}{c|l|c|cccc}
    \toprule
    Priv. Budget	 & Method   & \# Param. &  SST-2 & QNLI & QQP & MNLI \\
    \midrule
    \multirow{3}{*}{ $\epsilon = 6$ } & LoRA$_{r=8}$ & $1.57M$ &  $93.70$ & $84.99$  & $82.11$ & $39.46$ \\
    & FFA-LoRA$_{r=8}$ & $0.79M$ &  $93.73$ & $87.27$  & $83.31$ & $78.81$ \\
    &\textbf{FedTT}\cellcolor{blue!10} & 
     $\bm{0.24M}$\cellcolor{blue!10} &$\bm{93.80}$\cellcolor{blue!10} & $\bm{87.43}$\cellcolor{blue!10}  & $\bm{84.61}$ \cellcolor{blue!10}& $\bm{85.45}$\cellcolor{blue!10} 
    \cellcolor{blue!10} \\
    
    \midrule
    \multirow{3}{*}{ $\epsilon = 3$ } & LoRA$_{r=8}$ & $1.57M$ & $93.32$ & $83.94$  & $82.08$ & $35.82$ \\
    &FFA-LoRA$_{r=8}$ & $0.79M$ & $93.59$ & $86.18$  & $83.03$ & $77.42$ \\
    &\textbf{FedTT}\cellcolor{blue!10} &   $\bm{0.24M}$\cellcolor{blue!10}&$\bm{93.96}$ \cellcolor{blue!10}& $\bm{86.64}$  \cellcolor{blue!10}& $\bm{84.36}$ \cellcolor{blue!10}& $\bm{85.08}$ \cellcolor{blue!10}\\
    \midrule
    \multirow{3}{*}{ $\epsilon = 1$ }  & LoRA$_{r=8}$ & $1.57M$ &   $94.32$ & $88.95$  & $81.28$ & $33.80$  \\
    &FFA-LoRA$_{r=8}$   & $0.79M$ & $94.32$ & $90.35$  & $82.50$ & $75.05$  \\
    &\textbf{FedTT}\cellcolor{blue!10} &  
    $\bm{0.24M}$\cellcolor{blue!10} &$\bm{95.64}$\cellcolor{blue!10} & $\bm{91.67}$ \cellcolor{blue!10} & $\bm{83.11}$\cellcolor{blue!10} & $\bm{83.81}$\cellcolor{blue!10}  \\
    \bottomrule
    \end{tabular}}
\end{table}

We conduct experiments using the Generalized Language Understanding Evaluation (GLUE) benchmark~\cite{wang2018glue}, employing the complete training dataset for each task. We record the best validation results after the $100$ communication round for cross-silo FL and after the $1000$ communication round for LSCD FL. 
Initially, we compare FedTT and FedTT+ with various PEFT methods in a cross-silo FL setting, using the DeBERTa-base model and assuming independent and identically distributed (i.i.d.) data among clients. We set the number of clients to $5$ and the local training epochs to $1$. The results, shown in Table~\ref{table1}, indicate that FedTT achieves a higher average score compared to other methods with lower than $0.15M$ trainable parameters. Notably, FedTT exhibits only a $0.5\%$ performance gap compared to LoRA with a rank of $ 8 $, which has $5\times$ more trainable parameters. FedTT+ achieves comparable accuracy to other methods while drastically reducing the number of trainable parameters. It exhibits only a $0.6\%$ performance gap compared to LoRA with a rank of $4$, while using $6\times$ fewer trainable parameters, which significantly lowers communication overhead.

We then use the RoBERTa-base model to further evaluate FedTT against other PEFT methods. For the MRPC and SST-2 tasks, we employ a cross-silo FL setup with $10$ clients, while for the other tasks, we apply a LSCD FL configuration with $1000$ clients, randomly selecting $10$ clients per round. Our settings for the RoBERTa-base model align with those in~\cite{zhang2023fedpetuning}, allowing us to leverage their results of PEFT methods like LoRA, Adapter, Prefix, and BitFit under FL settings. Table~\ref{table2} shows that FedTT achieves a higher average score with the fewest trainable parameters among all methods. 
Additional details regarding Tables \ref{table1} and \ref{table2} are provided in Appendices \ref{a.3} and \ref{a.4}, respectively.

\subsection{Impact of Data Heterogeneity}\label{hetero}

\begin{figure}[t]
    \centering 
    \scriptsize
    \begin{subfigure}[b]{0.24\textwidth} 
        \centering
        \includegraphics[width=\textwidth]{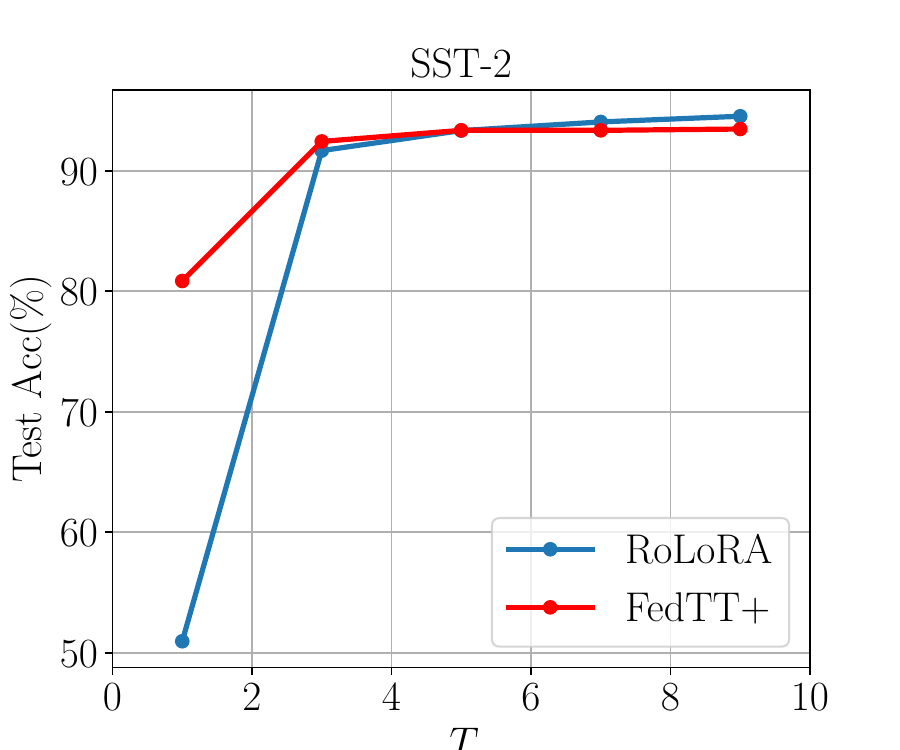}
    \end{subfigure}
    \hspace{-0.1cm} 
    \begin{subfigure}[b]{0.24\textwidth}
        \centering
        \includegraphics[width=\textwidth]{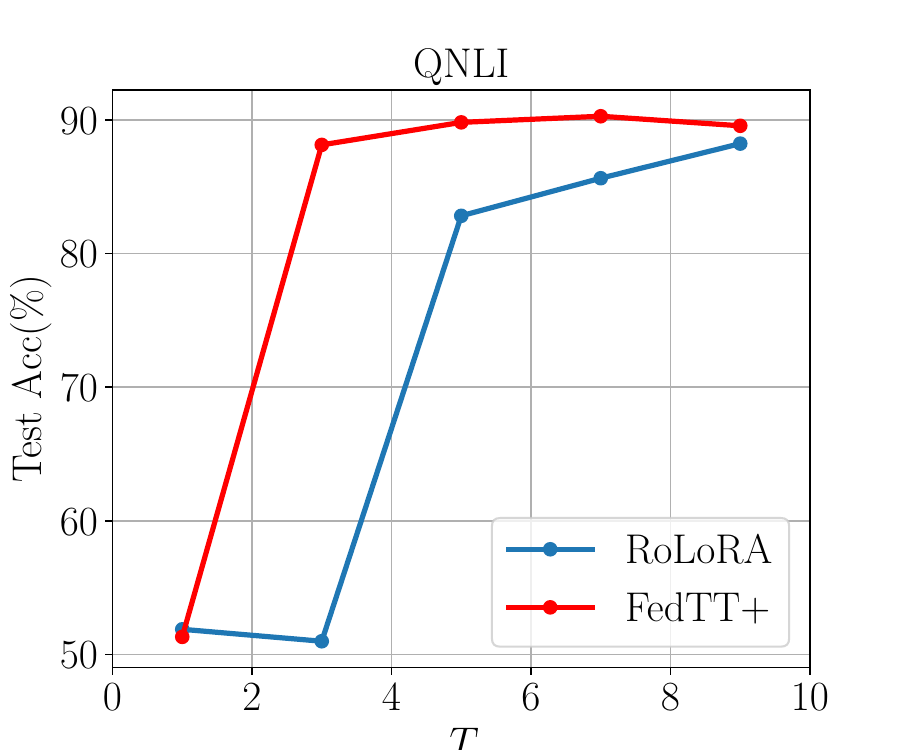}
    \end{subfigure}
    
    
    \begin{subfigure}[b]{0.24\textwidth}
        \centering
        \includegraphics[width=\textwidth]{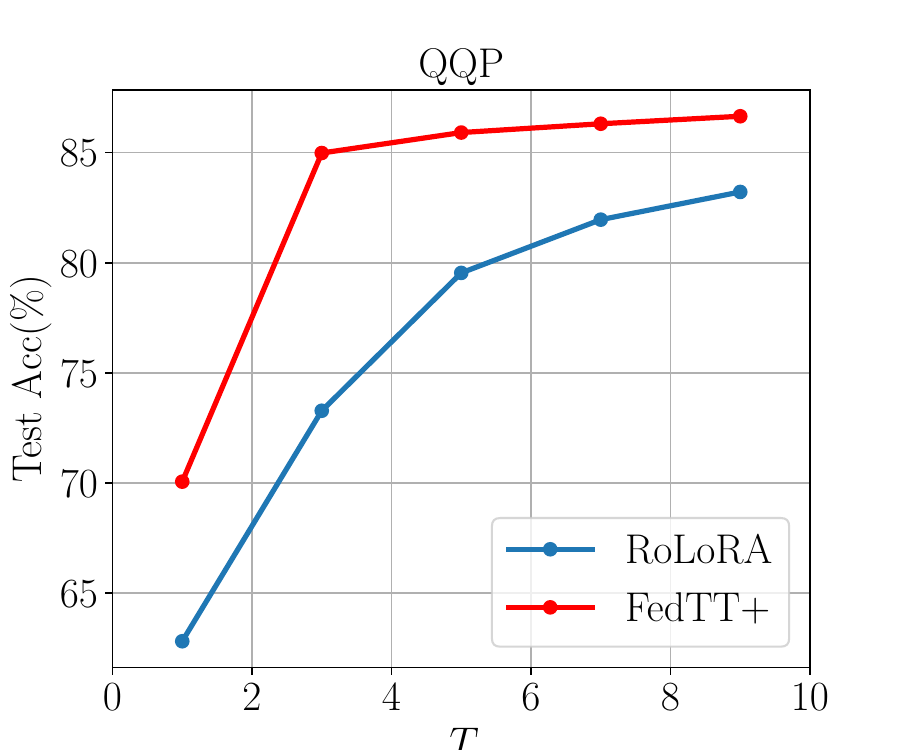}
    \end{subfigure}
    \hspace{-0.1cm}
    \begin{subfigure}[b]{0.24\textwidth}
        \centering
        \includegraphics[width=\textwidth]{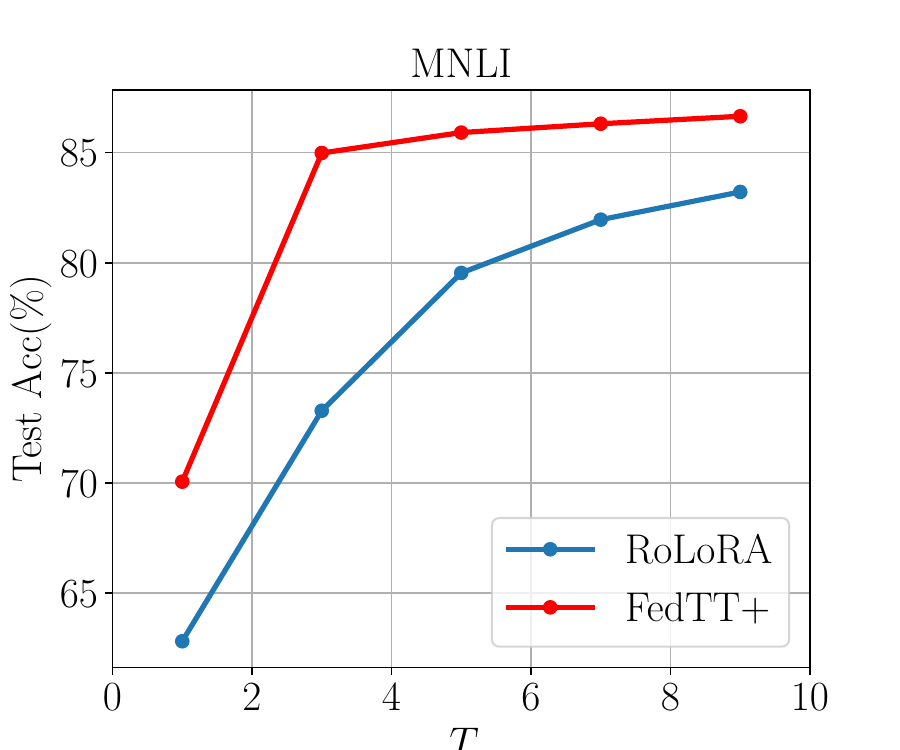}
    \end{subfigure}
    
    \caption{Test accuracies versus communication rounds for RoBERTa-Large models across different tasks in a cross-silo FL setting with 50 clients.}
    \label{fig2}
\end{figure}

\begin{table*}[t]
	\centering
	\caption{Comparative analysis of LoRA, FedTT, and FedTT+ using the LLaMA2-7B and LLaMA2-13B models.}
	\label{tab:LLaMA2_superglue}
	\resizebox{0.81\textwidth}{!}{%
		\begin{tabular}{c|lc|ccc|c}
			\toprule
    Fed. Set. 
			 &
    Model \& Method 
			 &
			\# Param. &
			\multicolumn{1}{c|}{COPA} &
			\multicolumn{1}{c|} {ReCoRD} &\multicolumn{1}{c|}{SQuAD} & Avg. \\ 
            
		 \midrule
		\multirow{2}{*}{ large-scale FL } &	LLaMA2-7B (LoRA$_{r=8}$)   & $4.19M$    & $87$   & $81.0$ & $90.49$ & $86.16$  \\ 
		&	\textbf{LLaMA2-7B (FedTT)}    & $0.52M$  & $90$  & $80.1$   & $89.32$ & $86.47$\\ 
   \midrule
\multirow{3}{*}{ cross-silo FL }&
   LLaMA2-13B (LoRA$_{r=8}$)              & $6.55M$ & $89$&	$83.6$&$91.07$&	 $87.89$   \\ 
	&		\textbf{LLaMA2-13B (FedTT)}      & $0.64M$ & $89$	& $83.5$&	$90.06$	&$87.52$   \\
   &\textbf{LLaMA2-13B (FedTT+)}    & $0.18M$  & $88$  & $83.7$   & $90.40$ &$87.70$\\
   
   \bottomrule
		\end{tabular}%
	}
 \label{table5}
\end{table*}

\begin{table}[t]
\caption{Total transmitted messages for the MNLI task.}
\centering
\adjustbox{width=1\linewidth}{
\begin{tabular}{@{}lccc@{}}
\toprule
Methods   & Total transmitted messages (KB)  & & Comm. overhead \\ \midrule
LoRA   & \phantom{11}1172 \progressbar{100}{black} & \hspace{-4.6cm}\textcolor{red}{\vline height 0.28cm depth 3pt width 1.5pt}& $\times$1.88\phantom{1} \\ 
BitFit      & \phantom{11}390 \hspace{0.2cm}\progressbar{31}{black} & \hspace{-4.6cm}\textcolor{red}{\vline height 0.28cm depth 3pt width 1.5pt}& $\times$0.62\phantom{1} \\
RoLoRA      & \phantom{11}624 \hspace{0.2cm}\progressbar{50}{black} & \hspace{-4.6cm}\textcolor{red}{\vline height 0.28cm depth 3pt width 1.5pt}& $\times$1.00\phantom{1} \\ 
Prompt      & \phantom{11}1523 \hspace{-0.2cm}$\;\;$\progressbar{100}{black} & \hspace{-4.6cm}\textcolor{red}{\vline height 0.28cm depth 3pt width 1.5pt}& $\times$2.44\phantom{1} \\ 
FedTT & \phantom{11}234 \hspace{0.2cm}\progressbar{19}{black} & \hspace{-4.6cm}\textcolor{red}{\vline height 0.28cm depth 3pt width 1.5pt}& $\times$0.37\phantom{1} \\ 
FedTT+ & \phantom{11}78\hspace{0.4cm}
\progressbar{6}{black} & \hspace{-4.6cm}\textcolor{red}{\vline height 0.28cm depth 3pt width 1.5pt}& $\times$0.12\phantom{1} \\ \bottomrule
\end{tabular}
}
\label{tab:comm}
\end{table}

Similar to~\cite{chenrobust}, we simulate varying levels of data heterogeneity by using different numbers of clients: $3$ clients (i.i.d. data distribution), $20$ clients (mild heterogeneity), and $50$ clients (severe heterogeneity), with $20$ local updates to further increase data heterogeneity. To ensure a comparable number of trainable parameters, we make the last $6$ layers trainable for the SST-2 dataset and the last $9$ layers for the other datasets in LoRA, FFA-LoRA, RoLoRA, and FedTT. In contrast, for FedTT+, all layers are made trainable. Additional details are provided in Appendix \ref{a.55}. Our RoBERTa-Large model settings align with those in~\cite{che2023federated}, allowing us to leverage their results for LoRA, FFA-LoRA, and RoLoRA methods. As shown in Table~\ref{table3}, FedTT+ consistently outperforms the other methods across varying heterogeneity settings while utilizing fewer parameters. This further demonstrates that, under a comparable number of trainable parameters and in the presence of data heterogeneity, FedTT+ outperforms FedTT, highlighting the effectiveness of adaptively freezing tensorized adapters in addressing data heterogeneity. 
 Fig. \ref{fig2} illustrates test accuracy versus the number of communication rounds in the severe heterogeneity scenario, showing that FedTT+ achieves faster convergence than RoLoRA across multiple tasks. An additional experiment examining the effect of data heterogeneity using an alternative heterogeneity measure is provided in Appendix \ref{hetero_app}.

\subsection{Performance on Larger Models}\label{app:large}
 We use SuperGLUE tasks~\cite{wang2019superglue} and a generation task (SQuAD~\cite{rajpurkar2016squad}) to compare FedTT and FedTT+ with LoRA. The results are shown in Table~\ref{table5}. We use the LLaMA2-7B model to simulate LSCD FL with $1000$ clients, where $10$ of them are chosen randomly in each round for training. Additional details are provided in Appendix \ref{a.5}. FedTT has performance nearly identical to LoRA with a rank of $8$, while achieving about $ 10\times $ lower communication overhead. For cross-silo FL, we utilize the LLaMA2-13B model with $ 10 $ clients. As seen in Table~\ref{table5}, both FedTT and FedTT+ demonstrate performance close to LoRA, with $ 10\times $ and $ 30\times $ lower communication overhead, respectively.
\subsection{Communication Cost Analysis}
In this section, we present the communication cost analysis for the MNLI task of Table \ref{table1} in Table \ref{tab:comm}. The analysis for all tasks of Table \ref{table1} and Table \ref{table3} can be found in Appendix \ref{comm_cost}. Following~\cite{guo2024selective}, we compare the proposed method with baselines in terms of:  
(i) Up-link message size (in KB) per communication round,  
(ii) Number of communication rounds required to reach $95$\% of the prediction accuracy reported in Table \ref{table1}, and  
(iii) Total size of transmitted messages. As demonstrated, our proposed algorithms effectively
minimize both the up-link message size and the total size of transmitted messages.

\subsection{Differential privacy guarantees}\label{diff:main}
We compare our proposed FedTT algorithm with baseline methods under differential privacy guarantees. Definitions and theoretical results related to DP are provided in Appendix \ref{privacy_app}. Following~\cite{sun2024improving}, we train the RoBERTa-Large model with a learning rate of $1e-3$ and three clients. We implement the DP-SGD algorithm using the Opacus~\cite{yousefpour2021opacus} library with privacy parameters $\delta = 1e-5$ and three different privacy budgets, $\epsilon \in \{1,3,6\}$. The clipping threshold is chosen from $C \in \{2,5\}$.  

We follow the experimental setup from~\cite{sun2024improving}, enabling direct evaluation against LoRA and FFA-LoRA. The results in Table~\ref{tab:priv} demonstrates that FedTT consistently achieves higher accuracy across different privacy budgets while using fewer trainable parameters. Notably, the largest performance gap is observed in the MNLI task, a three-class classification problem.

\subsection{Tensor Rank Analysis}
In this section, we evaluate the performance of the proposed FedTT method with varying tensor ranks across four datasets using the DeBERTa-base model. The number of clients is set to $5$, each performing one local update per round. We use a learning rate of \( 1e-3 \) and a batch size of $32$. The best validation results are reported after $20$ communication rounds. As shown in Table~\ref{table_ablation}, higher tensor ranks lead to better average accuracy, albeit at the cost of an increased number of trainable parameters. In our experiments, we set the rank to $5$, as it offers a favorable trade-off between accuracy and model size.

\begin{table}[t]
    \centering
    \caption{Tensor rank analysis using the DeBERTa-Base model.}
    \small
\resizebox{\linewidth}{!}{%
    \begin{tabular}{l|c|cccc|lrl}
    \toprule
    Model \& Method & \# Param.  &  SST-2 & QNLI & QQP & MNLI & Avg. \\
    \midrule
    FedTT$_{r=2}$ & $0.03M$ &  $93.58$ & $90.76$  & $86.79$ & $85.29$ & $89.10$ \\
    FedTT$_{r=5}$ & $0.06M$   &  $93.00$  & $91.87$ & $87.65$ & $85.91$ & $89.61$\\
    FedTT$_{r=10}$ & $0.17M$  &  $93.69$ & $92.48$  & $88.22$ & $87.01$ & $90.35$ \\
    \bottomrule
    \end{tabular}}
\label{table_ablation}
\end{table}

\section{Conclusion}
\vspace{-1pt}
This work presents a novel federated and tensorized fine-tuning framework, addressing the key challenges of communication overhead and data heterogeneity in FL. FedTT leverages tensorized adapters to significantly reduce communication costs while maintaining high performance, achieving up to a $10\times$ reduction in trainable parameters compared to existing methods. Additionally, FedTT+ enhances robustness in cross-silo FL by adaptively freezing portions of tensor factors, further optimizing parameter efficiency. Extensive experiments with models like BERT and LLaMA-2 demonstrate the effectiveness of FedTT and FedTT+ in both cross-silo and LSCD FL settings, offering a scalable and communication-efficient solution for fine-tuning LLMs in distributed environments.

\section*{Limitations}
In this work, while we addressed the data heterogeneity challenge, we did not conduct experiments on another significant challenge in FL: system heterogeneity. However, our framework has the potential to tackle this challenge as well, by allowing different tensor ranks to be assigned to clients based on their computational capabilities. This presents an interesting direction for future research.

\section*{Acknowledgments}
This work was supported by the National Science Foundation under Grant 2419982 and Grant 2342253. Z. Zhang and Y. Yang were supported by the U.S. Department of Energy, Office of Science, Office of Advanced Scientific Computing Research, Artificial Intelligence for Science program, under contract DE-SC0025390. This research used resources of the National Energy Research Scientific Computing Center, a DOE Office of Science User Facility supported by the Office of Science of the U.S. Department of Energy under Contract No. DE-AC02-05CH11231 using NERSC award ASCR-ERCAP0030039.

\clearpage
\bibliography{ref.bib}

\clearpage
\appendix


\section{Experiment setup}
\subsection{Dataset Setup}
We begin our experiments using the Generalized Language Understanding Evaluation (GLUE) benchmark~\cite{wang2018glue}, which includes a range of natural language understanding tasks. These tasks consist of perceptual analysis (SST2~\cite{socher2013recursive}), similarity and paraphrase tasks (MRPC, QQP~\cite{dagan2005pascal}), and natural language reasoning (MNLI, QNLI~\cite{williams2017broad, rajpurkar2018know}). The utilized metrics for the GLUE benchmark are summarized in Table \ref{tab:glue_metric}. We record the best validation results after $100$ communication rounds for cross-silo FL and after $1000$ communication rounds for large-scale cross-device FL. For the DeBERTa-Base models on the QQP task, we randomly select
$1000$ samples from the validation set and report the highest accuracy achieved.
\begin{table}[ht]
\centering
\caption{Dataset descriptions and statistics.}\label{tab:glue_metric}
\small
\begin{tabular}{l|c|c|c}
\toprule
\textbf{Task} & \textbf{\# Train} & \textbf{\# Dev.} & \textbf{Metric}  \\ \midrule
MRPC      & 3,301 &  367   & F1 Score \\ 
SST-2     & 66,675 &  674  & Accuracy \\
QNLI      & 103,695 &   5,463 & Accuracy \\
QQP       & 360,210 &   40,430 & Accuracy \\ 
MNLI   & 388,774 &   9,815 & Accuracy \\
\bottomrule
\end{tabular}
\end{table}
\\~\\
We then select two multiple-choice tasks (COPA and ReCoRD) from the SuperGLUE benchmark~\cite{wang2019superglue} and a question-answering generation task (SQuAD~\cite{rajpurkar2016squad}). The metrics used for evaluation are summarized in Table \ref{tab:superglue_metric}.

\begin{table}[ht]
	\centering
	\caption{The utilized metrics for the SuperGLUE benchmark.}
	\label{tab:superglue_metric}
	\resizebox{0.2\textwidth}{!}{%
		\begin{tabular}{@{}cc@{}}
			\toprule
			Task Name & Metric                       \\ \midrule
			COPA      & Accuracy                        \\
			ReCoRD      & F1                                    \\
			SQuAD    & F1                               \\
			\bottomrule
		\end{tabular}%
	}
\end{table}

\subsection{Additional Detail of TT-format}
In this paper, we use the TT format to structure the weight matrices within the tensorized layers. To accommodate matrices of varying shapes, we design specific tensor shapes based on the hidden sizes and bottleneck configurations of different models. The tensor shapes are outlined in Table~\ref{tab:shape}, with examples provided for the DeBERTa/RoBERTa-base and LLaMA-2-7b models, which have hidden sizes of 768 and 4096, respectively. For models with other hidden sizes, the appropriate tensor shape must be determined prior to training.
\begin{table}[!ht]
	\centering
	\caption{The shape settings of the TT-format}
	\label{tab:shape}
	\resizebox{0.45\textwidth}{!}{%
		\begin{tabular}{@{}ccc@{}}
			\toprule
			Modules & Matrix Shape & Tensor Shape\\ 
			\midrule
			Tensorized Adapters         & $768\times 64$      &   [8, 8, 12, 8, 8]    \\
			& $4096\times 64$   &   [16, 16, 16, 4, 4, 4]       \\
			& $64\times 768$      &   [8, 8, 12, 8, 8]    \\
			& $64\times 4096$   &   [4, 4, 4, 16, 16, 16]       \\
			\midrule
			Tenosrized Classifier (Optional)         &  $768\times 768$     &    [12, 8, 8, 8, 8, 12]    \\
			&  $768\times 768$     &    [8, 8, 8, 8, 8, 8, 8, 8]    \\
			\bottomrule
		\end{tabular}%
	}
\end{table}

\subsection{Additional Details for Table \ref{table1}} \label{a.3}
We use DeBERTa-Base models with the GLUE dataset in a cross-silo FL setup. The learning rate is selected from $[5e-3, 1e-3, 5e-4, 1e-4]$, and the batch size from $[16, 32]$ for all tasks and methods. The number of clients is set to $5$, with one local update per communication round. The best validation results were recorded after $100$ communication rounds. Additional details on the chosen parameters are provided in Table~\ref{tab:hyper_para}.
\begin{table}[ht]
	\centering
	\caption{ The hyperparameter grids used for GLUE experiments.}
	\label{tab:hyper_para}
	\resizebox{0.4\textwidth}{!}{%
		\begin{tabular}{@{}ccc@{}}
			\toprule
			Experiment & Hyperparameters & Values \\ \midrule
			LoRA   
			& Rank    &    ${4,8}$   \\ \midrule
RoLoRA   
			& Rank    &    $4$   \\ \midrule
			Bitfit         
			& Bias Terms    &   All    \\ \midrule
			Prompt     
			& \# Tokens   &     10   \\ \midrule
			P-tuning          
			& \# Tokens   &     20   \\ 
			&  Prompt Length   &     $[128,768]$  \\ \midrule
			FedTT         
			& Tensor Rank    &    $5$  \\
			\midrule
			FedTT+  
			& Tensor Rank    &    $5$   \\ \bottomrule
		\end{tabular}%
	}
\end{table}

\begin{table*}[ht]
    \centering
    \caption{Number of trainable parameters for results in Table~\ref{table3}. }

    \begin{tabular}{l|cccc|lrl}
    \toprule
    Model \& Method  &  SST-2 & QNLI & QQP & MNLI & Avg. \\
    \midrule
    RoBERTa-Large (LoRA$_{r=2}$)   &  $49K$ & $74K$  & $74K$ & $74K$ & $68K$ \\
    RoBERTa-Large (FFA-LoRA$_{r=2}$)  &  $\bm{25K}$ & $37K$  & $37K$ & $37K$ & $34K$ \\
    RoBERTa-Large (RoLoRA$_{r=2}$)   &  $\bm{25K}$ & $37K$  & $37K$ & $37K$ & $34K$ \\
    \textbf{RoBERTa-Large (FedTT)}  &  $39K$ & $55K$  & $55K$ & $55K$ & $51K$ \\
    \textbf{RoBERTa-Large (FedTT+)}  &  $28K$ & $\bm{28K}$  & $\bm{28K}$ & $\bm{28K}$ & $\bm{28K}$ \\
    \bottomrule
    \end{tabular}
\label{table4}
\end{table*}

            
   

\subsection{Additional Details for Table \ref{table2}} \label{a.4}
We follow the experimental setup of~\cite{zhang2023fedpetuning}, using RoBERTa-Base models with the GLUE dataset in both cross-silo and large-scale cross-device FL configurations. The learning rate was selected from $[1e-2, 5e-3, 1e-3, 5e-4, 1e-4, 5e-5]$, and batch size from $[16, 32]$ across all tasks and methods. For the MRPC and SST-2 tasks, we use a cross-silo FL setup with $10$ clients, while for the other tasks, we employ a large-scale cross-device FL setup with $1000$ clients, selecting $10$ clients per round. Accuracy results for Prefix, Adapter, LoRA, and BitFit methods are sourced from~\cite{zhang2023fedpetuning}. In addition to reporting the accuracy of the FedTT method, we also includ the P-Tuning method with prompt lengths $[128, 768]$, and IA3~\cite{liu2022few} method, to further enrich the experiments.

\subsection{Additional Details for Table \ref{table3}}\label{a.55}
    
    
    
We adopt the same settings as~\cite{chenrobust}, using RoBERTa-Large models with the GLUE dataset in cross-silo FL scenarios. To simulate varying levels of data heterogeneity, we report the accuracy of our method for $3$ clients (i.i.d. data distribution), $20$ clients (moderate heterogeneity), and $50$ clients (high heterogeneity), with $20$ local updates. The learning rate is set to $1e-3$ for all tasks, with a batch size of $64$ for SST-2 and $32$ for other tasks. To ensure a comparable number of trainable parameters, we make the last $6$ layers trainable for the SST-2 dataset and the last $9$ layers for the other datasets in LoRA, FFA-LoRA, RoLoRA, and FedTT. In contrast, for FedTT+, all layers are made trainable.
 We provide a comparison of the number of trainable parameters for LoRA, FFA-LoRA, RoLoRA, and our proposed FedTT, and FedTT+ methods in Table~\ref{table4}. Accuracy results for LoRA, FFA-LoRA, and RoLoRA are sourced from~\cite{chenrobust}. 


\subsection{Additional Details for Table \ref{tab:LLaMA2_superglue}}\label{a.5}
Due to the large number of parameters in LLaMA-2, it is rarely used in FL scenarios, and few existing works provide results for such models. However, FedTT significantly reduces communication overhead, making it feasible to utilize these models in FL settings. We evaluate FedTT and FedTT+ against LoRA using SuperGLUE tasks~\cite{wang2019superglue} and a generation task (SQuAD~\cite{rajpurkar2016squad}). The results are presented in Table~\ref{table5}.  
\\~\\
For each task, we randomly select $1000$ samples for training and $1000$ for validation, reporting the best validation accuracy. We use a learning rate of $1e-4$, batch size of $2$, and $3$ local updates across all tasks. In large-scale cross-silo FL, we simulate training with the LLaMA2-7B model across $1000$ clients, selecting $10$ randomly per round. FedTT achieves performance nearly identical to LoRA with a rank of $8$, while reducing communication overhead by approximately $10\times$.  
\\~\\
For cross-silo FL, we utilize the LLaMA2-13B model with $10$ clients. As shown in Table~\ref{table5}, both FedTT and FedTT+ achieve performance close to LoRA while reducing communication overhead by approximately $10\times$ and $30\times$, respectively.

\section{Additional Experiment on Data Heterogeneity}\label{hetero_app}
\begin{table*}[t]
    \centering
    \caption{Comparison of cross-silo FL methods using RoBERTa-Large models under varying degrees of data heterogeneity. The accuracies for LoRA and FFA-LoRA methods are sourced from~\cite{sun2024improving}.}\label{hetero2}
    \small

    \begin{tabular}{c|l|c|cccc|lrl}
    \toprule
    Data Dist. & Model \& Method  & \# Param. &  SST-2 & QNLI & QQP & MNLI & Avg. \\
    \midrule
    \multirow{3}{*}{ i.i.d. } & RoBERTa-Large (LoRA$_{r=8}$) & $1.57M$ &  $94.42$ & $91.38$  & $84.47$ & $86.90$ & $89.29$ \\
    &RoBERTa-Large (FFA-LoRA$_{r=8}$) & $0.79M$ &  $95.14$ & $92.64$  & $86.31$ & $87.13$ & $90.30$ \\
    &\textbf{RoBERTa-Large (FedTT+)}\cellcolor{blue!10} & $\bm{0.03M}$\cellcolor{blue!10} &  $\bm{95.41}$\cellcolor{blue!10} & $\bm{94.05}$\cellcolor{blue!10}  & $\bm{88.15}$ \cellcolor{blue!10}& $\bm{88.65}$\cellcolor{blue!10} & $\bm{91.56}$
    \cellcolor{blue!10} \\
    
    \midrule
    \multirow{3}{*}{ mild het. } & RoBERTa-Large (LoRA$_{r=8}$) &  $1.57M$ &  $93.55$ & $91.36$  & $84.41$ & $87.01$ & $89.08$ \\
    &RoBERTa-Large (FFA-LoRA$_{r=8}$) & $0.79M$ &  $94.10$ & $91.62$  & $85.33$ & $87.04$ & $89.52$ \\
    &\textbf{RoBERTa-Large (FedTT+)}\cellcolor{blue!10} & $\bm{0.03M}$\cellcolor{blue!10} &  $\bm{95.53}$ \cellcolor{blue!10}& $\bm{93.54}$  \cellcolor{blue!10}& $\bm{87.91}$ \cellcolor{blue!10}& $\bm{88.45}$ \cellcolor{blue!10}& $\bm{91.36}$ \cellcolor{blue!10}\\
    \midrule
    \multirow{3}{*}{ sever het. }  & RoBERTa-Large (LoRA$_{r=8}$) &  $1.57M$ &  $94.32$ & $88.95$  & $83.51$ & $82.03$ & $87.20$ \\
    &RoBERTa-Large (FFA-LoRA$_{r=8}$) & $0.79M$  &  $94.32$ & $90.35$  & $84.35$ & $85.05$ & $88.52$ \\
    &\textbf{RoBERTa-Large (FedTT+)}\cellcolor{blue!10} & $\bm{0.03M}$\cellcolor{blue!10}  &  $\bm{95.64}$\cellcolor{blue!10} & $\bm{91.67}$ \cellcolor{blue!10} & $\bm{86.66}$\cellcolor{blue!10} & $\bm{87.73}$\cellcolor{blue!10} & $\bm{90.42}$\cellcolor{blue!10} \\
    \bottomrule
    \end{tabular}
\label{tab:hetero_new}
\end{table*}
Data heterogeneity is a common challenge in most practical scenarios~\cite{hajihashemimulti}. In FL, local models on individual clients can diverge from the global model’s optimal state, resulting in slower convergence~\cite{hsieh2020non,li2020federated}. This issue is particularly pronounced when training LLMs in federated environments, as data heterogeneity can severely impact model performance~\cite{zhang2023fedpetuning}. 
\\~\\
As mentioned before, three primary factors contribute to data heterogeneity: (1) the number of clients, (2) the number of local updates, and (3) non-i.i.d. data distribution. In Section \ref{hetero} of the paper, which evaluates the heterogeneity robustness of FedTT+, we considered the effects of the number of clients and local updates. Specifically, we increased the number of clients to intensify heterogeneity.
In this section,  we conduct an additional experiment to evaluate the performance of FedTT+ under varying levels of data heterogeneity. Similar to~\cite{sun2024improving}, we train the RoBERTa-Large model and consider three levels of data distribution for three clients:
\begin{itemize}  
    \item \textbf{i.i.d. Data Distribution}: Data is evenly distributed across all clients.  
    \item \textbf{Mild Heterogeneity}: For binary classification tasks, data is split as $[0.15, 0.85]$, $[0.85, 0.15]$, and $[0.5, 0.5]$ among three clients. For three-class classification tasks, the splits are $[0.6, 0.2, 0.2]$, $[0.2, 0.6, 0.2]$, and $[0.2, 0.2, 0.6]$.  
    \item \textbf{Severe Heterogeneity}: For binary classification tasks, data is split as $[0.05, 0.95]$, $[0.95, 0.05]$, and $[0.5, 0.5]$. For three-class classification tasks, the splits are $[0.9, 0.05, 0.05]$, $[0.05, 0.9, 0.05]$, and $[0.05, 0.05, 0.9]$.  
\end{itemize}
To further amplify heterogeneity, we used $10$ local updates. We followed the exact experimental setup described in~\cite{sun2024improving}, allowing us to directly compare our results with those reported for LoRA and FFA-LoRA. The results are presented in Table~\ref{hetero2}.
As shown in the Table, FedTT+ consistently outperforms LoRA and FFA-LoRA across different heterogeneity settings while utilizing fewer parameters.

\begin{table*}[ht]
    \centering
    \caption{Communication cost analysis for Table \ref{table1}.}\label{comm1}
    \small
\begin{tabular}{c|c|cccc|cccc}
\toprule 
\multirow[t]{2}{*}{Method} & \multirow[t]{2}{*}{Up-link Message Size (KB)} & \multicolumn{4}{|c|}{\# Communication Round} & \multicolumn{4}{|c}{Total transmitted messages (KB)} \\
\midrule &  & SST-2 & QNLI & QQP & MNLI & SST-2 & QNLI & QQP & MNLI \\
\midrule LoRA$_{r=4}$  & $586$ & $2$ & $2$ & $2$ & $2$ & $1172$ & $1172$ & $1172$ & $1172$  \\
 RoLoRA$_{r=4}$ & $312$ & $2$ & $2$ & $1$ & $2$ & $624$ & $624$ & $312$ & $624$\\
 FedTT & $234$ & $2$ & $2$ & $2$ & $2$& $468$ & $\bm{468}$ & $468$ & $468$ \\
 FedTT+ & $\bm{78}$ & $5$ & $6$ & $3$ & $3$ & $\bm{390}$ & $\bm{468}$ & $\bm{234}$ & $\bm{234}$\\
\bottomrule
\end{tabular}
\end{table*}

\begin{table*}[ht]
    \centering
    \caption{Communication cost analysis for sever heterogeneity in Table \ref{table3}.}\label{comm2}
    \small
\begin{tabular}{c|c|cccc|cccc}
\toprule
\multirow[t]{2}{*}{Method} & \multirow[t]{2}{*}{Up-link Message Size (KB)} & \multicolumn{4}{|c|}{\# Communication Round} & \multicolumn{4}{|c}{Total transmitted messages (KB)} \\
\midrule &  & SST-2 & QNLI & QQP & MNLI & SST-2 & QNLI & QQP & MNLI \\
\midrule RoLoRA$_{r=2}$ & $133$ & $3$ & $10$ & $11$ & $11$ & $399$ & $1330$ & $1463$ & $1463$\\
 FedTT & $199$ & $3$ & $3$ & $3$ & $3$ & $597$ & $597$ & $597$ & $597$ \\
 FedTT+ & $\bm{109}$ & $3$ & $3$ & $3$ & $3$ & $\bm{327}$ & $\bm{327}$ & $\bm{327}$ &$ \bm{327}$\\
\bottomrule
\end{tabular}
\end{table*}

\section{Communication Cost Analysis}\label{comm_cost}
In this section, we provide communication cost analysis for Table \ref{table1} and Table \ref{table3}. Specifically, we compare the proposed method with baselines in terms of communication cost, following \cite{guo2024selective}. The analysis includes:
\begin{itemize}
    \item Up-link message size (in KB) for each communication round.
    \item 
Number of communication rounds needed to reach the predefined target performance on the SST-2, QNLI, QQP, and MNLI datasets.
\item 
Total transmitted messages (in KB).
\end{itemize}
The target performance is defined as $95$\% of the prediction accuracy reported in Table \ref{table1} and Table \ref{table3}.  
\\~\\
In a federated learning system, two key parameters influence communication efficiency: the up-link message size and the total size of transmitted messages, computed as the product of the number of communication rounds and the up-link message size. We provide detailed comparisons in Table \ref{comm1} and Table \ref{comm2}. As demonstrated, our proposed algorithms effectively minimize both the up-link message size and the total number of transmitted messages.

\section{Privacy}\label{privacy_app}
While LLMs excel in performance due to their transformer-based architecture and vast parameter count, their ability to memorize and inadvertently reveal sensitive information from training data is a concern \cite{carlini2021extracting,rathod2025privacy}. A widely adopted framework for mitigating such privacy risks is DP \cite{dwork2006calibrating}, which provides formal guarantees against data leakage.  
\\~\\
In this section, we first outline key definitions of DP and introduce the DP-SGD algorithm. We then establish a theoretical privacy guarantee within the DP framework. 
\\~\\
\textbf{Definition 1.} ($(\epsilon, \delta)$-Differential Privacy \cite{dwork2006calibrating}])
A randomized mechanism $\mathcal{M}: \mathcal{D} \to \mathcal{R}$ satisfies $(\epsilon, \delta)$-differential privacy if for any two adjacent datasets $D, D' \in \mathcal{D}$ that differ in at most one data point, and for any subset of possible outputs $S \subseteq \mathcal{R}$, the following holds:
\begin{align}
    \Pr[\mathcal{M}(D) \in S] \leq e^{\epsilon} \Pr[\mathcal{M}(D') \in S] + \delta.\nonumber
\end{align}
DP ensures that the mechanism $\mathcal{M}$ provides privacy guarantees by limiting the impact of any single data point on the output, with $\epsilon$ controlling the privacy loss and $\delta$ allowing for a small probability of failure in the guarantee.
\\~\\
\textbf{Differentially Private Stochastic Gradient Descent (DP-SGD) Algorithm~\cite{abadi2016deep}:}
DP-SGD is a modification of the SGD algorithm designed to provide differential privacy guarantees. It achieves this by introducing two key modifications:
\begin{itemize}
    \item \textbf{Gradient Clipping:} To ensure that individual data points do not have a disproportionate influence on the model update, each per-sample gradient is clipped to a fixed norm $C$.
    \item \textbf{Noise Addition:} After aggregating the clipped gradients, Gaussian noise $z \sim \mathcal{N}\left(0, C^2 \sigma^2 I\right)$ is added to the sum of clipped gradients in a batch $\mathcal{B}$ from the dataset $\mathcal{D}$ before updating the model parameters. This helps obscure the contribution of any single data point.
\end{itemize}
The noisy sum of clipped gradients is computed as:

\begin{equation}
\bar{g} = \frac{\sum_{i \in \mathcal{B}} \operatorname{Clip}(\nabla f_i, C) + z}{|\mathcal{B}|},\nonumber
\end{equation}
which is then used to update the model. Here, the noise scale $\sigma$ is determined based on sequential composition rules, given the privacy parameters $\epsilon$ and $\delta$, the number of iterations $T$, and the sampling probability $q = |\mathcal{B}|/|\mathcal{D}|$.
\\~\\
Although other techniques like secure aggregation have been explored in privacy-preserving FL \cite{behnia2024efficient}, our focus here is on DP. In federated learning with DP, the level of privacy protection depends on whether the aggregation server is trusted by the clients. In global DP, the server is trusted, allowing clients to send raw model updates, while the server applies differential privacy by adding noise to the aggregated updates before releasing them~\cite{mcmahan2017learning}. In contrast, local DP assumes an untrusted server, requiring each client to add noise to their updates before transmission, ensuring privacy even if the server is compromised~\cite{wu2020value}. In Section \ref{diff:main}, we adopt the stronger local DP approach to guarantee robust privacy protection without relying on a trusted server.  
\\~\\
\textbf{Proposition 1.}
The mechanism for updating FedTT using locally run DP-SGD and FedAvg satisfies $(\epsilon, \delta)$-DP, given that  
$\forall i$, the sampling probability is defined as $q=\left|\mathcal{B}_i\right|/\left|\mathcal{D}_i\right|$.  
Furthermore, the total number of local updates per client is denoted as $T$, and $\sigma$ is chosen as  
\begin{align}
    \sigma = O\left(\frac{q \sqrt{T \log (1 / \delta)}}{\epsilon}\right).\nonumber
\end{align}
\textit{Proof.} Similar to \cite{sun2024improving}.

\end{document}